\documentclass[letterpaper, 10 pt, conference]{ieeeconf}  

\IEEEoverridecommandlockouts                              

\usepackage{times}

\usepackage{algorithm}
\usepackage[noend]{algcompatible}

\usepackage{amsmath}
\usepackage{amsfonts}
\usepackage{breqn}
\usepackage{multirow,hhline,graphicx,array}
\graphicspath{ {./images/} }
\usepackage{multicol}
\usepackage[bookmarks=true]{hyperref}
\usepackage{subfiles}
\usepackage{caption}
\usepackage{float}
\usepackage[flushleft]{threeparttable}

\usepackage{xcolor}

\title{\LARGE \bf
RELLIS-3D Dataset: Data, Benchmarks and Analysis
}

\author{Peng Jiang$^{1}$, Philip Osteen$^{2}$, Maggie Wigness$^{2}$ and Srikanth Saripalli$^{1}$
\thanks{$^{1}$J. Mike Walker '66 Department of Mechanical Engineering, 
Texas A\&M University, College Station, TX 77843, USA
        {\tt\small \{maskjp, ssaripalli\}@tamu.edu}}%
\thanks{$^{2}$DEVCOM Army Research Laboratory (ARL), Adelphi, MD 20783, USA
        {\tt\small \{philip.r.osteen, maggie.b.wigness\}.civ@mail.mil}}%
}

\begin{document}

\maketitle
\thispagestyle{empty}
\pagestyle{empty}

\begin{abstract}
Semantic scene understanding is crucial for robust and safe autonomous navigation, particularly so in off-road environments. Recent deep learning advances for 3D semantic segmentation rely heavily on large sets of training data, however existing autonomy datasets either represent urban environments or lack multimodal off-road data. We fill this gap with RELLIS-3D, a multimodal dataset collected in an off-road environment, which contains annotations for 13,556 LiDAR scans and 6,235 images. The data was collected on the Rellis Campus of Texas A\&M University, and presents challenges to existing algorithms related to class imbalance and environmental topography. Additionally, we evaluate the current state of the art deep learning semantic segmentation models on this dataset. Experimental results show that RELLIS-3D presents challenges for algorithms designed for segmentation in urban environments. This novel dataset provides the resources needed by researchers to continue to develop more advanced algorithms and investigate new research directions to enhance autonomous navigation in off-road environments. RELLIS-3D is available at \url{https://github.com/unmannedlab/RELLIS-3D}
\end{abstract}

\section{Introduction}
Autonomous navigation systems that rely solely on LiDAR and an inertial navigation system (INS) have been shown to perform poorly in off-road environments \cite{Maturana2018, Jackel2006, Thrun2006}. These systems leverage geometric information but lack higher-level semantic understanding of the environment that could make path planning and navigation more efficient. For example, bushes are identified as obstacles by LiDAR, but for larger platforms, these parts of the environment may actually be traversable. The importance of semantic awareness has led to the emergence of visual perception systems that directly feed information to navigation systems to complement depth sensors \cite{Maturana2018, Hussein2016, Huertas2005, Manduchi2005, Yang2020}. LiDAR provides highly accurate 3D information, is not affected by varying illumination, and adds reflectance information to characterize object surfaces uniquely, whereas cameras provide dense color and texture information to obtain fine-grained semantic information.

Semantic segmentation for indoor and urban scenes has made great advancements using the numerous large-scale datasets available in these domains~\cite{Maturana2018, Feng2020,Brostow2009, Cordts2016, Caesar2019,Silberman2012, Zhou2017}. 
Compared with urban road scenes and indoor environments, off-road environments have unstructured class boundaries, uneven terrain, strong textures, and irregular features that preclude the direct transfer of models between the different types of environments. Also, there are large differences in class distributions across distinct off-road environments. Although there are some off-road datasets available \cite{Maturana2018, RUGD2019IROS, valada16iser}, the autonomous navigation research community still lacks a multi-modal dataset with a large number of ground truth annotations for developing reliable autonomous robot systems in off-road environments.

\begin{figure}
  \centering
  \includegraphics[width=0.5\textwidth]{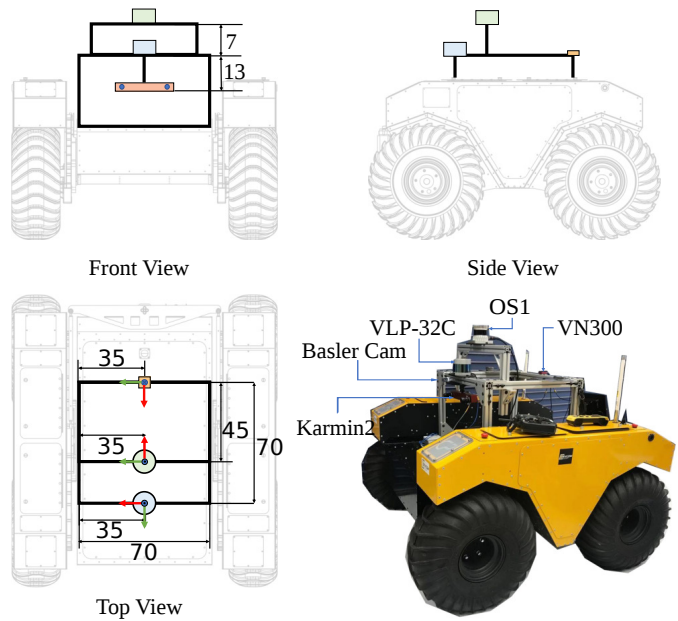}
  \caption{Warthog Platform Configuration. Illustration of the dimensions and mounting positions of the sensors with respect to the robot body. (Units: cm) }
  \label{fig:sensor_setup}
\end{figure}
\vspace{-5pt}

To address the need for multi-modal data resources to advance autonomous navigation in off-road environments, we present RELLIS-3D, a novel dataset captured from a Clearpath Robotics Warthog platform (shown in Fig.~\ref{fig:sensor_setup}). All data were recorded in the Ground Research facility on the Rellis Campus of Texas A\&M University, which is an off-road environment that includes different runways, aprons, terrain, forested areas, bushes, pastures, and lakes.
\begin{table*}[h]
\centering
\vspace{10pt}
\captionsetup{margin={0pt,0pt}}
\caption{Overview of off-road datasets. Ours is by far the largest dataset with multiple modalities.}
\begin{tabular}{c|c|c|c|c|c}
\hline
Name             & Sensors       & \# Annotations\textsuperscript{1}        & \# Classs\textsuperscript{2} & Annotation Type      & Modality                        \\
\hline
RUGD \cite{RUGD2019IROS}            & camera        & 7546       & 24     & pixel-wise       & RGB                             \\
\hline
DeepScene \cite{valada16iser}        & camera        & 366        & 6      & pixel-wise       & RGB, Depth, NIR, NRG, NDVI, EVI \\
\hline
Pezzementi et al \cite{Pezzementi2018} & camera        & 95000      & 1      & bounding box     & RGB                             \\
\hline
YCOR \cite{Maturana2018}             & camera        & 1076       & 8      & pixel-wise       & RGB                             \\
\hline
Dabbiru et al \cite{Dabbiru2020}    & simLiDAR      & 2743       & 6      & point-wise       & Point Cloud                     \\
\hline
\textbf{Ours}             & camera, LiDAR & 6235/13556 & 20     & pixel/point-wise & RGB, Point Cloud                \\
\hline
\end{tabular}

\begin{tablenotes}
\centering
  \small
  \item  \textsuperscript{1} Number of images/scans annotated, \textsuperscript{2}Number of classes annotated.
\end{tablenotes}
\label{tbl:dataset_tabel}
\vspace{-10pt}

\end{table*}

The RELLIS-3D dataset is comprised of a large set of raw sensor data synchronized with Precision Time Protocol (PTP), including color camera images, laser scans, high-precision global positioning measurements, inertial measurement from a combined Global Positioning and Inertial Navigation System (GPS/INS), and depth images from a 3D stereo camera. By including the full set of raw autonomy data, we facilitate additional algorithms, such as those that fuse visual and inertial/depth data, to be developed and tested without any new data collection.

Beyond the raw sensor data, RELLIS-3D will be released with ground truth annotations for a subset of the LiDAR scans and RGB camera images. To the best of the authors' knowledge, RELLIS-3D represents the first multi-modal off-road navigation dataset with synchronized raw sensor data and a large number of ground truth annotations. The major contributions of this dataset can be summarized as follows:
\begin{itemize}
  \item We release five sequences of synchronized sensor data captured while driving in off-road environments in Robot Operating System (ROS) bag format, including RGB camera images, LiDAR point clouds, a pair of stereo images, high-precision GPS measurement, and IMU data.
  \item Across the five sequences of collected data we provide 6,235 pixel-wise image annotations, and semantic labels for 13,556 full LiDAR point cloud scans.
  \item We establish a benchmark by defining training, validation, and testing splits, and provide an initial analysis using state-of-the-art image and point cloud semantic segmentation algorithms. These results demonstrate the challenges of semantic segmentation of off-road data and help identify open areas of research that RELLIS-3D can help advance.
\end{itemize}

\section{Related Work}
Several datasets have been published in the last decade for scene understanding research for autonomous vehicles and robots. A large majority of existing data represent urban environments with mostly on-road navigation. Of these datasets, most only include 2D annotations (e.g., bounding boxes or region masks) for RGB camera images. Examples include CamVid~\cite{Brostow2009}, Cityscapes~\cite{Cordts2016}, Mapillary Vistas~\cite{Neuhold2017}, D2-City~\cite{Che2019}, and BDD100k~\cite{Yu2018}. However, RGB cameras are not the only sensors used in autonomous driving, therefore several multi-modal datasets have also been published, e.g., KITTI \cite{Geiger2013IJRR}, nuScenes\cite{Caesar2019} and A2D2~\cite{Geyer2020}. And the SemanticKITTI \cite{behley2019iccv} further enrich KITTI by adding a large amount of semantic annotation for its LiDAR subset. 

For off-road autonomous navigation research to mirror the progress made for operating in urban environments, high quality data resources must be made available to the research community. Yet, there are comparatively far fewer off-road datasets than those from urban environments. Table \ref{tbl:dataset_tabel} outlines the existing off-road datasets available and their data and annotation characteristics. RUGD~\cite{RUGD2019IROS} is an RGB image dataset for semantic segmentation in off-road environments with a rich ontology and large set of ground truth annotations but lacks multiple modalities. The Freiburg Forest dataset \cite{valada16iser} provides multi-modal, multi-spectral image data
but lacks additional sensor modalities, e.g., LiDAR, and only 366 images have ground truth annotations. YCOR \cite{Maturana2018} provides both image and point cloud data, but only images are annotated. NREC Agricultural Person-Detection Dataset is a dataset for person detection in off-road environments, but only provides bounding box annotations for a single class. Dabbiru et al. \cite{Dabbiru2020,Goodin2018} present a framework that generates simulated labeled point cloud data and trains a Convolution Neural Network (CNN) for LiDAR semantic segmentation, which could provide a route for sim2real transfer learning with real data such as RELLIS-3D. There is a clear gap in the available data for off-road navigation and RELLIS-3D fills this gap by providing a full stack of multi-modal sensor data and multi-modal annotation with a rich ontology. 
\section{Sensor Setup and Calibration}
\subsection{Sensors}
Our sensor setup is illustrated in Fig.\ref{fig:sensor_setup} and includes:

\begin{itemize}
  \item 1 \(\times\) Ouster OS1 LiDAR: 64 Channels, 2048 horizontal resolution, 10 Hz, 45$^{\circ}$ vertical field of view
  \item 1 \(\times\) Velodyne Ultra Puck: 32 Channels, 10hz, 40$^{\circ}$ vertical field of view
  \item 1 \(\times\) Nerian Karmin2 + Nerian  SceneScan: 3D Stereo Camera, 10 hz
  \item 1 \(\times\) RGB Camera: Basler acA1920-50gc camera with 16mm/F18 EDMUND Optics lens, image resolution 1920x1200, 10 hz
  \item Inertial Navigation System (GPS/IMU):  Vectornav VN-300 Dual Antenna GNSS/INS, 300 Hz GPS, 100 Hz IMU
\end{itemize}
In addition to the sensor suite, our Warthog platform has two computers. The navigation computer is responsible for the robotic control, while the vision computer is devoted to data collection and sensor processing. The two computers communicate through Ethernet connection and are synchronized using PTP. Both computers run Ubuntu Linux (64 bit) and ROS Kinetic to collect the incoming data streams. 

\subsection{Synchronization}
We use PTP to synchronize the sensors throughout the computer and sensor network. The vision computer is synchronized with the navigation computer, Ouster LiDAR, Stereo Camera, and RGB Camera. Unfortunately, the GPS/IMU system cannot be synchronized with PTP, however the GPS/IMU system provides updates at 100 Hz. In this case, we complete the synchronization by chosing the GPS/IMU information with the closest timestamp to the LiDAR and camera timestamp for a particular frame.
\begin{figure*}
\vspace{5pt}
  \centering
  \includegraphics[]{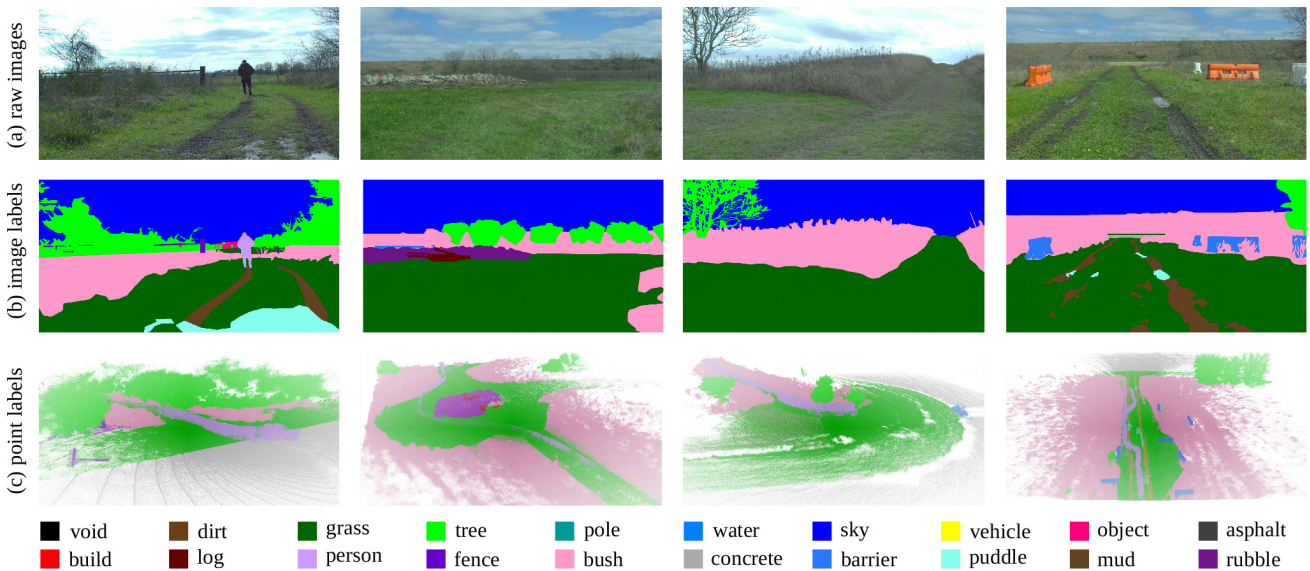}
  \caption{Ground truth annotations examples provided in the RELLIS-3D dataset. Images are densely annotated with pixel-wise labels from 20 different visual classes. LiDAR scans are point-wise labeled with the same ontogloy.}
  \label{fig:data_exp}
\end{figure*}
\subsection{Camera Calibration}
We assume our cameras fit a pin-hole projection model in which a 3D scene is projected on an image plane using a perspective transform. In addition to the standard projection parameters expressed by a calibration matrix $\textbf{K}$, real lenses also exhibit radial and tangential distortions given by distortion coefficients contained in a vector $\textbf{D}$. For this dataset, we use the \texttt{ROS Camera Calibrator}\footnote{\url{http://wiki.ros.org/camera_calibration}} library for intrinsic camera calibration. This package is a ROS wrapper for the camera calibration functionality provided in OpenCV \cite{opencvlibrary}. 

\subsection{LiDAR \& Camera Calibration}
The knowledge of the extrinsic calibration between sensors is of paramount importance for fusing information from all sensing modalities. Most perception and state estimation algorithms assume the extrinsic calibration to be known \emph{a-priori}. We determine the extrinsic calibration between cameras and 3D-LiDARs using \cite{PBPC,mishra2020experimental,osteen_2105}.

As described in the aforementioned works, in the absence of ground truth, we verify our algorithms by comparing the estimated parameters against the given factory stereo calibration and by projecting points lying on the edges of the planar target in the LiDAR frame on the camera image and calculating the mean line re-projection errors (MLRE). MLRE is an independent evaluation metric since the calibration algorithms described in \cite{PBPC} and \cite{mishra2020experimental} do not use it as a residual in their respective optimizations.

\section{Dataset}
\subsection{Data Description}
RELLIS-3D includes five traversal sequences, and was collected on three non-paved trails of the Ground Research facility on the Rellis Campus of Texas A\&M University. Three sequences were recorded on the first trail that was covered with bushes and sparse trees. These sequences differ in the direction the robot is moving on the trail, and the day the data was collected. Another sequence captured the environment of the second trail that passes a pasture and traverses a forested area. The last sequence is recorded on a hill that is surrounded by a lake and highway. Sequences were collected by teleoperating the robot to follow the trail, and each sequence includes around fives minutes of data.

With the goal of providing multi-modal data to enhance autonomous off-road navigation, we defined an ontology of object and terrain classes, which largely derives from the RUGD dataset \cite{RUGD2019IROS} but also includes unique terrain and object classes not present in RUGD. Specifically, sequences from this dataset includes classes such as mud, man-made barriers, and rubble piles. Additionally, this dataset provides a finer-grained class structure for water sources, i.e., puddle and deep water, as these two classes present different traversability scenarios for most robotic platforms. Overall, 20 classes (including void class) are present in the data. The full ontology can be seen in Fig.~\ref{fig:data_exp}. 
\subsection{Annotations}
Pixel-wise image annotations were provided by Appen\footnote{https://appen.com/}, a company that leverages crowdsourcing for training data annotation to be used for a variety of AI/ML tasks. To ensure annotation consistency, work was assigned to trained annotators, and a single annotator was assigned a sequence of frames from a single video sequence. Annotations from the crowdsourcing platform also underwent several rounds of in-house verification to correct missing, incorrect, or inconsistent labels. We downsample the camera stream to 5Hz for ground truth labeling, and do not duplicate annotations for which the robot is stationary, resulting in a total of $6,235$ images with pixel-wise annotations.

Point-wise annotation for 3D point clouds is initialized by using the camera-LiDAR calibration to project the more than $6,000$ image annotations onto point clouds. Using the 3D point cloud annotation application provided by SemanticKITTI\cite{behley2019iccv}, annotations are refined for the multiple overlapped LiDAR scans. LiDAR scan alignment is crucial to obtain quality annotations and although our system has a highly accurate INS there are still map inconsistencies. To address this issue, we first register and loop close the sequences using a SLAM system~\cite{ Maturana2015} and output each scan's position based on the SLAM results. Using this process the $13,556$ scans received full point-wise annotations, where each scan includes up to $13,056$ points.
\subsection{Dataset Statistics}
\begin{figure}
  \centering
  \includegraphics[]{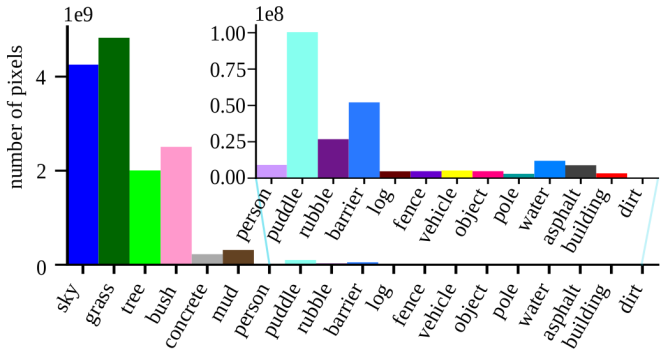}
  \caption{Image Label distribution. The sky, grass, tree and bush constitute the major classes.}
  \label{fig:img_dist}
\end{figure}

\begin{figure}
  \centering
  \includegraphics[]{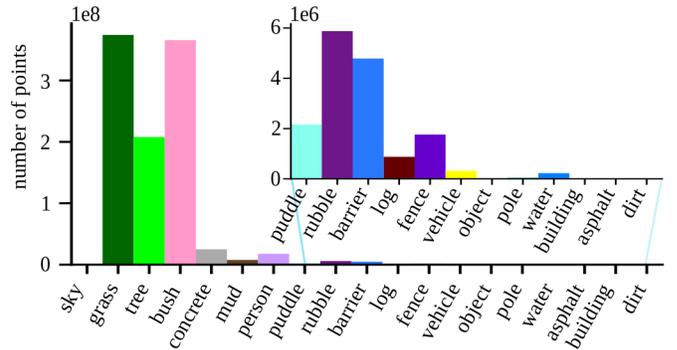}
  \caption{Point Cloud Label distribution. The grass, tree, and bush also dominate the population.}
  \label{fig:pt_dist}
\end{figure}

Figures \ref{fig:img_dist} and \ref{fig:pt_dist} show the class distribution breakdown for image and point cloud annotations, respectively. The class distribution among both modalities is highly imbalanced. For image annotations, sky, grass, tree, and bushes make up 94\% of the total labeled pixels. Among the LiDAR data, grass, tree, and bushes make up 80\% of the total point labels. Differences in resolution, viewing angles, and sensor mechanism leads to the divergence in label distributions between the image and point cloud data. For example, because of the sensor mechanism, LiDAR is unable to detect sky, but because the LiDAR has a 360$^{\circ}$ viewing angle it picks up more person labels than the imagery since human operators usually followed the robot during data collection.

The non-uniform class distribution present in RELLIS-3D is common among datasets used for semantic segmentation \cite{Feng2020, Neuhold2017, behley2019iccv}. Moreover, class imbalance is a problem that perception algorithms will encounter upon deployment.
Although imbalanced class distributions exist in almost all current available urban datasets, the overall class distributions across distinct urban datasets are quite similar \cite{behley2019iccv, Neuhold2017}. However, the class imbalances in off-road environments are highly dependent on the particular environment, and the class imbalance within a dataset is more severe, making the semantic segmentation of rare classes more challenging than for urban environments. 
\section{Benchmarks, Evaluation Metrics and Experiments}

\begin{table*}[]
    \vspace{10pt}
    \captionsetup{margin={0pt,0pt}}
    \caption{Single image (20 classes) / scan (16 classes) for all baselines on test set.}
    \footnotesize  
    \setlength{\tabcolsep}{2.5pt}
    \renewcommand{\arraystretch}{1}
    \centering
    
    \begin{tabular}{ c | c  c  c  c  c  c c  c  c  c  c  c  c c  c c c c | c } 
    \hline
    models  & {\rotatebox[origin=c]{90}{sky}} & {\rotatebox[origin=c]{90}{grass}}  & {\rotatebox[origin=c]{90}{tree}}  & {\rotatebox[origin=c]{90}{bush}} & {\rotatebox[origin=c]{90}{concrete}} & {\rotatebox[origin=c]{90}{mud}}  & {\rotatebox[origin=c]{90}{person}}    & {\rotatebox[origin=c]{90}{puddle}}     & {\rotatebox[origin=c]{90}{rubble}}    & {\rotatebox[origin=c]{90}{barrier}}    & {\rotatebox[origin=c]{90}{log}}        & {\rotatebox[origin=c]{90}{fence}}     & {\rotatebox[origin=c]{90}{vehicle}}      & {\rotatebox[origin=c]{90}{object}}       & {\rotatebox[origin=c]{90}{pole}}   & {\rotatebox[origin=c]{90}{water}}    & {\rotatebox[origin=c]{90}{asphalt}}     & {\rotatebox[origin=c]{90}{building}}       & {\rotatebox[origin=c]{90}{mean}}\\ 
    \hline
    hrnet+OCR & 96.94 & 90.20 & 80.53 & 76.76 & 84.22 & 43.29 & 89.48 & 73.94 & 62.03 & 54.86 & 0.00 & 39.52 & 41.54 & 46.44 & 9.51 & 0.72 & 33.25 & 4.60  & 51.55 \\ 
    gscnn & 97.02 & 84.95 & 78.52 & 70.33 & 83.82 & 45.52 & 90.31 & 71.49 & 66.03 & 55.12 & 2.92 & 41.86 & 46.51 & 54.64 & 6.90 & 0.94 & 44.18 & 11.47  & 52.92 \\
    \hline \hline
    salsanext& - &  64.74 & 79.04 & 72.90 & 75.27 & 9.58 & 83.17 & 23.20 & 5.01 & 75.89 & 18.76 & 16.13& 23.12 & -  & 56.26 & 0.00 & - & -   & 43.07 \\
    kpconv& - & 56.41 & 49.25 & 58.45 & 33.91 & 0.00 & 81.20 & 0.00 & 0.00 & 0.00 & 0.00 & 0.40 & 0.00 & - & 0.00 & 0.00 & - & -   & 19.97 \\
    \hline  
    \end{tabular}
    
    \label{tbl:mious_table}
    \vspace{-5pt}
    \end{table*}
\subsection{Evaluation Metrics}
For semantic segmentation, we evaluate algorithm performance with the widely used mean intersection-over-union (mIoU) metric \cite{Everingham2014}, given by
\begin{equation}
    mIoU=\frac{1}{C}\sum_{c=1}^{C}\frac{TP_c}{TP_c+FP_c+FN_c}
\end{equation}
where \(TP_c\), \(FP_c\) and \(FN_c\) represent the number of true positive, false positive and false negative predictions for class \(c\), and \(C\) is the number of classes.

\subsection{Image Semantic Segmentation}
We provide an evaluation of 2D image semantic segmentation on our dataset using two state-of-the-art architectures: HRNETV2+OCR \cite{WangSCJDZLMTWLX19, Yuan2019} and Gated-SCNN \cite{Takikawa2019}.

 HRNETv2+OCR uses the High-Resolution Network (HRNet)\cite{WangSCJDZLMTWLX19} as its backbone and the Object-Contextual Representations (OCR) model \cite{Yuan2019} to explore the object-contextual representation of each pixel. HRNet maintains high-resolution representations throughout the whole model, unlike most other backbones that downsample input first and then upsample the features. OCR improves the pixel representation by aggregating the pixel features lying in the object region. 

Gated-SCNN\cite{Takikawa2019} is a two-stream CNN architecture for semantic segmentation that explicitly processes shape information in a separate branch, yielding a classical segmentation stream and a parallel shape stream. The architecture uses the higher-level activation in the classical stream to predict semantic segmentation and the low-level activation in the shape stream to abstract the shape information.

For this experiment, we split RELLIS-3D into a training set with 3,302 images, a validation set with 983 images, and a testing set with 1,672 images. When creating these splits, we tried to keep the training set large, while creating a testing set that was diverse including both similar and dissimilar scenarios seen in the training set. The image experiment uses 19 classes\footnote{We omit the dirt class in this evaluation because it is extremely sparse in the annotations.} (including void) for training and testing.

\begin{figure*}
\vspace{5pt}
  \centering
  \includegraphics[]{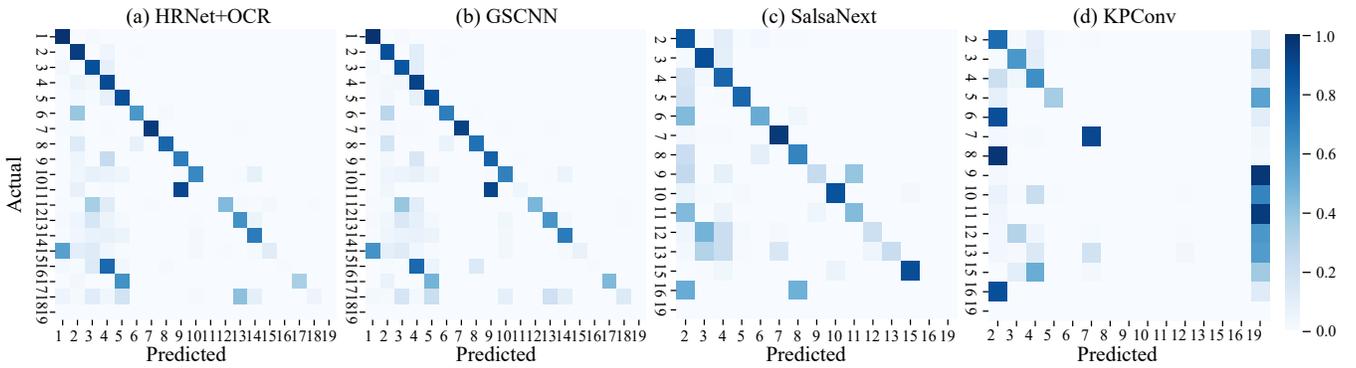}
  \caption{Confusion matrix. The y and x axis numbers represent classes ids (1: sky; 2: grass; 3: tree; 4: bush; 5: concrete; 6: mud; 7: person; 8: puddle; 9: rubble; 10: barrier; 11: log; 12: fence; 13: vehicle; 14: object; 15: pole; 16: water; 17: asphalt;18: building; 19: void.) Note that the sky label is omitted for the point cloud algorithm confusion matrices.}   
  \label{fig:cm}
\end{figure*}

Table \ref{tbl:mious_table} and Fig. \ref{fig:cm} shows the results of the image segmentation evaluation, which fall short of expectation. The Gated-SCNN model achieved only 52.92\% mIoU, while it reached 74.7\% mIoU \cite{Takikawa2019} on Cityscapes. HRNet+OCR achieved 51.55\% mIoU, but reached 81.1\% mIoU on Cityscapes. We believe the performance degradation is mainly caused by our dataset's serious class imbalance. This is supported by comparing Fig. \ref{fig:img_dist} and Fig. \ref{fig:cm}. Notice that the classes log, pole, water, and building have the lowest IoUs, and also represent the classes with fewest labels. These incorrect detections are of importance as they affect navigation decisions; for example, water is not traversable, unlike puddle, and human-made poles in an off-road area might provide warning information. Beyond the imbalance, the off-road environment's unclear boundaries also cause problems for both algorithms and human labelers. For humans, the indefinite boundaries make annotation difficult as compared with an urban roadway. The GSCNN algorithm utilizes boundary information to help perform segmentation, but this might lead to performance degradation for classes such as bush, grass, and trees with unclear boundaries (see Table \ref{tbl:mious_table}). For these classes, perfect boundary segmentation might not be necessary, but this problem inspires us to design new algorithms that can focus on the boundaries of specific classes, such as human-made objects and water.

\subsection{LiDAR Point Cloud Semantic Segmentation}
There are two types of deep learning methods for LiDAR Point Cloud: point-based methods and the projective method \cite{Guo2020}. We provide an evaluation of point cloud semantic segmentation on our dataset using two state-of-the-art architectures: SalsaNext~\cite{Cortinhal2020} and KPConv~\cite{Thomas2019}.

SalsaNext is an uncertainty-aware semantic segmentation model for full 3D LiDAR point cloud in real-time. SalsaNext has an encoder-decoder architecture and works on projected LiDAR point cloud data. The encoder introduces a new residual dilated convolution stack with gradually increasing receptive fields, while the decoder uses the pixel-shuffle layer to recover the resolution instead of deconvolution or upsampling layers.

Kernel Point Convolution (KPConv) is a specified, designed point cloud convolution operation that is more flexible than fixed grid convolution. KPConv uses a series of local, 3D convolution kernels to apply to the input points close to them, using a k-d tree to find nearby points. The regular subsampling strategy in the paper makes the KPconv operation more efficient and robust to varying densities. In the experiment, we use the KP-FCNN architecture \cite{Thomas2019} for semantic segmentation.

For the point cloud experiment, we follow the same data splits as for the image data. The training set has 7,800 scans, the validation set has 2,413 scans, and the testing set has 3,343 scans. Because the LiDAR scans are unable to establish points for classes such as sky, or objects far away (e.g., buildings in RELLIS-3D), the point cloud experiments use only the 15 classes with annotations (including void) for training and testing. 

Table \ref{tbl:mious_table} and Fig. \ref{fig:cm} show the results of the point cloud semantic segmentation. SalsaNext achieved 43.07\% mIoU and KPConv achieved 19.07\% mIoU, which is far under their performance on SemanticKITTI dataset, which was 59.5\% mIoU and 58.8\%, respectively. The imbalance phenomena in the point cloud dataset challenges these algorithms as well.
Compared with SalsaNext, the degradation of KPConv is more obvious. We believe that the extremely imbalanced and unstructured features of our dataset mainly cause the degradation. While training, the KPConv model does not learn on the whole LiDAR scan but instead on sampled neighborhoods of selected points. During training, there are two sampling strategies: random sampling and sampling based on the label distribution. The second strategy tried to mitigate the imbalanced distribution problem, but this attempt only increases results by \textbf{0.6\%} mIoU.

\subsection{Discussion} 
  \begin{figure}[h]
  \centering
    \begin{tabular}{cc}
     \includegraphics[width=0.2\textwidth]{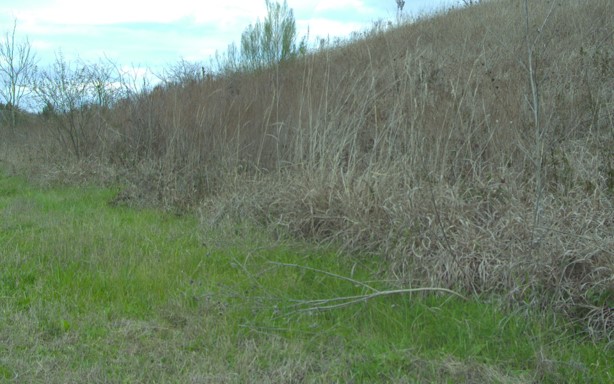} &   
     \includegraphics[width=0.2\textwidth]{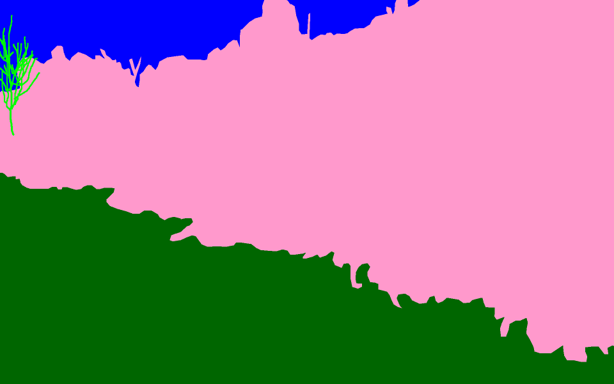} \\
      (a) &  
      (b) \\
     \includegraphics[width=0.2\textwidth]{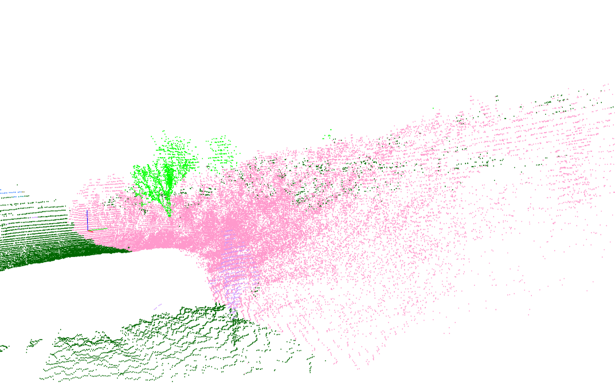} &   
     \includegraphics[width=0.2\textwidth]{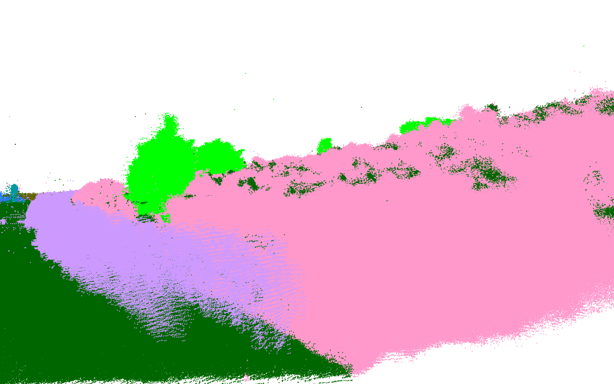} 
      \\
      (c) & (d)  \\
    \end{tabular}
    \caption{Hill Example (a) RGB Image, (b) Image Label, (c) Single Scan, (d) Multiple Scans}    
    \label{fig:hill_ex}
\end{figure}

  \begin{figure}[h]
  \centering
    \begin{tabular}{cc}
     \includegraphics[width=0.2\textwidth]{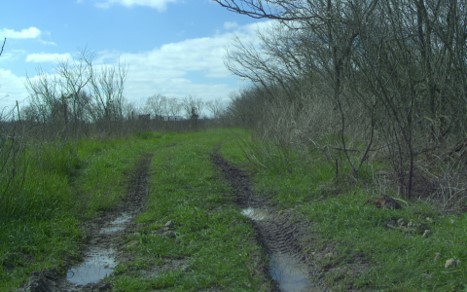} &   
     \includegraphics[width=0.2\textwidth]{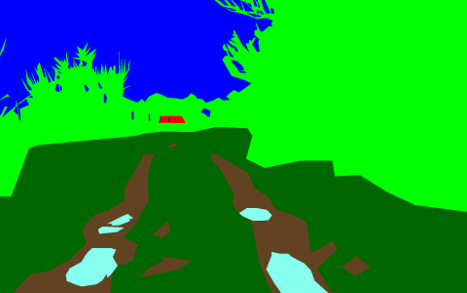} \\
      (a) &  
      (b) \\
     \includegraphics[width=0.2\textwidth]{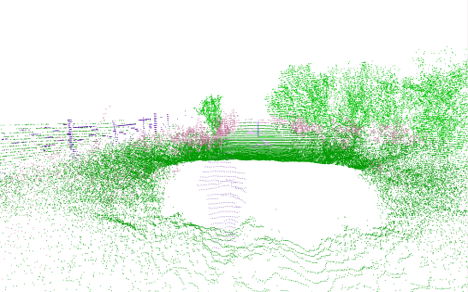} &   
     \includegraphics[width=0.2\textwidth]{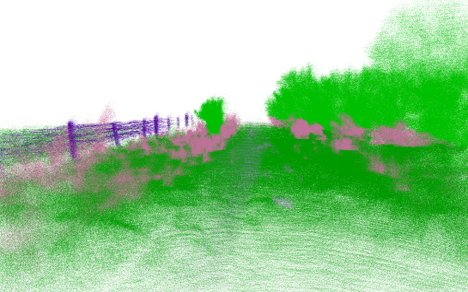} \\
      (c) &  
      (d) \\
    \end{tabular}
    \caption{Fence Example (a) RGB Image, (b) Image Label, (c) Single Scan, (d) Registered Scans. Note the clear difference in visibility of the fence class (shown in purple) between image and point clouds.}    
    \label{fig:fence_ex}
\end{figure}
\vspace{-5pt}
From the experimental results, we see that extreme class imbalance in off-road environments presents challenges for current semantic segmentation algorithms. For off-road environments tasks, the objects that are only visible in a small number of frames can be important for autonomous decision making and navigation planning. New algorithms might consider knowledge transfer from on-road datasets and fusion between two environments. Compared with the LiDAR point cloud, RGB images are more dense and contain more visually rich features to learn from, resulting in better performance for the image-based models on the semantic segmentation task.
However, LiDAR has its own advantages and we provide examples of why this modality is critical to pair with RGB data for autonomous navigation. Figure \ref{fig:hill_ex} shows an example frame from a sequence with a hill that is covered by bushes. Using both point cloud and image semantic understanding, the robot could recognize bushes successfully and decide whether to navigate among them, but a wrong navigation decision might be made without considering the 3D information. Using geometry-based methods alone, the bushes might be classified as non-traversable obstacle, but using semantic information a traversal decision could be made using higher level autonomy components. This example inspires us to consider new possible geometry-based ontology definitions, to complement existing semantic defintions. Currently, semantic labels are defined on the classes of objects but do not contain more abstract semantic information such as traversable/non-traversable, or up/down-hill. While some geometric supplemental information can be directly measured or inferred from measurement, some a-priori properties (e.g., traversable vs non-traversable) could prove beneficial for autonomous navigation tasks. Regardless, we believe that inertial as well as visual and depth information are vital for autonomous off-road tasks, and that the different data types complement each other.
For example, Figure \ref{fig:fence_ex} shows an example where a fence was behind bushes and very difficult to detect with image data or even single LiDAR scans. By registering scans over a small time window, the integrated LiDAR scans can successfully detect the fence, which could prevent incorrect or even hazardous navigation decisions.

\section{Summary and Future Work}
We introduce RELLIS-3D, a large-scale multimodal dataset with pixel-wise and point-wise annotations representative of off-road environments for semantic segmentation. The dataset presents challenges related to imbalanced class distributions, and unstructured features common in off-road environments. We show that performance of current state-of-the-art deep learning models degrades significantly on RELLIS-3D, indicating the uniqueness of off-road navigation and the need to further semantic segmentation research in these unstructured environments.  
We plan to extend RELLIS-3D in future work by including higher-order semantic labels such as traversable, non-traversable, up-hill and down-hill. Since the data was collected all in a single outdoor facility, maintaining diversity across the train/test/val set is a challenge. As we expand the dataset to different offroad environments, the environmental diversity will be greatly improved. In addition,  as we provide high accurate GPS data and stereo images in our datasets, we will investigate adding odometry benchmarks for RELLIS-3D and explore using semantic information to improve the odometry estimation results.






\bibliographystyle{IEEEtran}
\bibliography{IEEEabrv,references}

\begin{thebibliography}{10}
\providecommand{\url}[1]{#1}
\csname url@rmstyle\endcsname
\providecommand{\newblock}{\relax}
\providecommand{\bibinfo}[2]{#2}
\providecommand\BIBentrySTDinterwordspacing{\spaceskip=0pt\relax}
\providecommand\BIBentryALTinterwordstretchfactor{4}
\providecommand\BIBentryALTinterwordspacing{\spaceskip=\fontdimen2\font plus
\BIBentryALTinterwordstretchfactor\fontdimen3\font minus
  \fontdimen4\font\relax}
\providecommand\BIBforeignlanguage[2]{{%
\expandafter\ifx\csname l@#1\endcsname\relax
\typeout{** WARNING: IEEEtran.bst: No hyphenation pattern has been}%
\typeout{** loaded for the language `#1'. Using the pattern for}%
\typeout{** the default language instead.}%
\else
\language=\csname l@#1\endcsname
\fi
#2}}

\bibitem{Maturana2018}
D.~Maturana, P.-W. Chou, M.~Uenoyama, and S.~Scherer, ``{Real-Time Semantic
  Mapping for Autonomous Off-Road Navigation},'' in \emph{F. Serv. Robot.},
  M.~Hutter and R.~Siegwart, Eds.\hskip 1em plus 0.5em minus 0.4em\relax Cham:
  Springer International Publishing, 2018, pp. 335--350.

\bibitem{Jackel2006}
\BIBentryALTinterwordspacing
L.~D. Jackel, E.~Krotkov, M.~Perschbacher, J.~Pippine, and C.~Sullivan, ``{The
  DARPA LAGR program: Goals, challenges, methodology, and phase I results},''
  \emph{Journal of Field Robotics}, vol.~23, no. 11-12, pp. 945--973, nov 2006.
  [Online]. Available: \url{http://doi.wiley.com/10.1002/rob.20161}
\BIBentrySTDinterwordspacing

\bibitem{Thrun2006}
\BIBentryALTinterwordspacing
S.~Thrun, ``{Winning the DARPA grand challenge},'' in \emph{Lecture Notes in
  Computer Science (including subseries Lecture Notes in Artificial
  Intelligence and Lecture Notes in Bioinformatics)}, vol. 4213 LNAI,
  no.~4.\hskip 1em plus 0.5em minus 0.4em\relax Springer Verlag, 2006, p.~4.
  [Online]. Available:
  \url{https://link.springer.com/chapter/10.1007/11871842{\_}4}
\BIBentrySTDinterwordspacing

\bibitem{Hussein2016}
A.~Hussein, P.~Marin-Plaza, D.~Martin, A.~{De La Escalera}, and J.~M. Armingol,
  ``{Autonomous off-road navigation using stereo-vision and laser-rangefinder
  fusion for outdoor obstacles detection},'' in \emph{IEEE Intelligent Vehicles
  Symposium, Proceedings}, vol. 2016-Augus.\hskip 1em plus 0.5em minus
  0.4em\relax Institute of Electrical and Electronics Engineers Inc., aug 2016,
  pp. 104--109.

\bibitem{Huertas2005}
A.~Huertas, L.~Matthies, and A.~Rankin, ``{Stereo-based tree traversability
  analysis for autonomous off-road navigation},'' in \emph{Proceedings -
  Seventh IEEE Workshop on Applications of Computer Vision, WACV 2005}.\hskip
  1em plus 0.5em minus 0.4em\relax IEEE Computer Society, 2005, pp. 210--217.

\bibitem{Manduchi2005}
\BIBentryALTinterwordspacing
R.~Manduchi, A.~Castano, A.~Talukder, and L.~Matthies, ``{Obstacle detection
  and terrain classification for autonomous off-road navigation},''
  \emph{Autonomous Robots}, vol.~18, no.~1, pp. 81--102, jan 2005. [Online].
  Available: \url{http://www.gao.gov/new.items/d01311.pdf}
\BIBentrySTDinterwordspacing

\bibitem{Yang2020}
\BIBentryALTinterwordspacing
Y.~Yang, D.~Tang, D.~Wang, W.~Song, J.~Wang, and M.~Fu, ``{Multi-camera visual
  SLAM for off-road navigation},'' \emph{Robotics and Autonomous Systems}, vol.
  128, p. 103505, jun 2020. [Online]. Available:
  \url{https://linkinghub.elsevier.com/retrieve/pii/S0921889019308711}
\BIBentrySTDinterwordspacing

\bibitem{Feng2020}
D.~Feng, C.~Haase-Schutz, L.~Rosenbaum, H.~Hertlein, C.~Glaser, F.~Timm,
  W.~Wiesbeck, and K.~Dietmayer, ``{Deep Multi-Modal Object Detection and
  Semantic Segmentation for Autonomous Driving: Datasets, Methods, and
  Challenges},'' \emph{IEEE Transactions on Intelligent Transportation
  Systems}, pp. 1--20, feb 2020.

\bibitem{Brostow2009}
\BIBentryALTinterwordspacing
G.~J. Brostow, J.~Fauqueur, and R.~Cipolla, ``{Semantic object classes in
  video: A high-definition ground truth database},'' \emph{Pattern Recognit.
  Lett.}, vol.~30, no.~2, pp. 88--97, 2009. [Online]. Available:
  \url{http://mi.eng.cam.ac.uk/research/projects/VideoRec/}
\BIBentrySTDinterwordspacing

\bibitem{Cordts2016}
\BIBentryALTinterwordspacing
M.~Cordts, M.~Omran, S.~Ramos, T.~Rehfeld, M.~Enzweiler, R.~Benenson,
  U.~Franke, S.~Roth, and B.~Schiele, ``{The Cityscapes Dataset for Semantic
  Urban Scene Understanding},'' in \emph{Proc. IEEE Comput. Soc. Conf. Comput.
  Vis. Pattern Recognit.}, vol. 2016-December.\hskip 1em plus 0.5em minus
  0.4em\relax IEEE Computer Society, dec 2016, pp. 3213--3223. [Online].
  Available: \url{http://arxiv.org/abs/1604.01685}
\BIBentrySTDinterwordspacing

\bibitem{Caesar2019}
\BIBentryALTinterwordspacing
H.~Caesar, V.~Bankiti, A.~H. Lang, S.~Vora, V.~E. Liong, Q.~Xu, A.~Krishnan,
  Y.~Pan, G.~Baldan, and O.~Beijbom, ``{nuScenes: A multimodal dataset for
  autonomous driving},'' mar 2019. [Online]. Available:
  \url{http://arxiv.org/abs/1903.11027}
\BIBentrySTDinterwordspacing

\bibitem{Silberman2012}
P.~K. Nathan~Silberman, Derek~Hoiem and R.~Fergus, ``Indoor segmentation and
  support inference from rgbd images,'' in \emph{ECCV}, 2012.

\bibitem{Zhou2017}
\BIBentryALTinterwordspacing
B.~Zhou, H.~Zhao, X.~Puig, S.~Fidler, A.~Barriuso, and A.~Torralba, ``{Scene
  parsing through ADE20K dataset},'' in \emph{Proceedings - 30th IEEE
  Conference on Computer Vision and Pattern Recognition, CVPR 2017}, vol.
  2017-Janua, 2017, pp. 5122--5130. [Online]. Available:
  \url{http://groups.csail.mit.edu/vision/datasets/ADE20K/}
\BIBentrySTDinterwordspacing

\bibitem{RUGD2019IROS}
M.~Wigness, S.~Eum, J.~G. Rogers, D.~Han, and H.~Kwon, ``A rugd dataset for
  autonomous navigation and visual perception in unstructured outdoor
  environments,'' in \emph{IROS}, 2019.

\bibitem{valada16iser}
A.~Valada, G.~Oliveira, T.~Brox, and W.~Burgard, ``Deep multispectral semantic
  scene understanding of forested environments using multimodal fusion,'' in
  \emph{ISER}, 2016.

\bibitem{Pezzementi2018}
\BIBentryALTinterwordspacing
Z.~Pezzementi, T.~Tabor, P.~Hu, J.~K. Chang, D.~Ramanan, C.~Wellington, B.~P.
  {Wisely Babu}, and H.~Herman, ``{Comparing apples and oranges: Off-road
  pedestrian detection on the National Robotics Engineering Center agricultural
  person-detection dataset},'' \emph{J. F. Robot.}, vol.~35, no.~4, pp.
  545--563, jun 2018. [Online]. Available:
  \url{http://doi.wiley.com/10.1002/rob.21760}
\BIBentrySTDinterwordspacing

\bibitem{Dabbiru2020}
L.~Dabbiru, C.~Goodin, N.~Scherrer, and D.~Carruth, ``{LiDAR Data Segmentation
  in Off-Road Environment Using Convolutional Neural Networks (CNN)},'' in
  \emph{SAE Tech. Pap.}, vol. 2020-April, no. April.\hskip 1em plus 0.5em minus
  0.4em\relax SAE International, apr 2020.

\bibitem{Neuhold2017}
G.~Neuhold, T.~Ollmann, S.~R. Bulo, and P.~Kontschieder, ``{The Mapillary
  Vistas Dataset for Semantic Understanding of Street Scenes},'' in \emph{Proc.
  IEEE Int. Conf. Comput. Vis.}, vol. 2017-October.\hskip 1em plus 0.5em minus
  0.4em\relax Institute of Electrical and Electronics Engineers Inc., dec 2017,
  pp. 5000--5009.

\bibitem{Che2019}
\BIBentryALTinterwordspacing
Z.~Che, G.~Li, T.~Li, B.~Jiang, X.~Shi, X.~Zhang, Y.~Lu, G.~Wu, Y.~Liu, and
  J.~Ye, ``{D{\$}{\^{}}2{\$}-City: A Large-Scale Dashcam Video Dataset of
  Diverse Traffic Scenarios},'' Tech. Rep., 2019. [Online]. Available:
  \url{http://arxiv.org/abs/1904.01975}
\BIBentrySTDinterwordspacing

\bibitem{Yu2018}
\BIBentryALTinterwordspacing
F.~Yu, H.~Chen, X.~Wang, W.~Xian, Y.~Chen, F.~Liu, V.~Madhavan, and T.~Darrell,
  ``{BDD100K: A Diverse Driving Dataset for Heterogeneous Multitask
  Learning},'' may 2018. [Online]. Available:
  \url{http://arxiv.org/abs/1805.04687}
\BIBentrySTDinterwordspacing

\bibitem{Geiger2013IJRR}
A.~Geiger, P.~Lenz, C.~Stiller, and R.~Urtasun, ``Vision meets robotics: The
  kitti dataset,'' \emph{International Journal of Robotics Research (IJRR)},
  2013.

\bibitem{Geyer2020}
\BIBentryALTinterwordspacing
J.~Geyer, Y.~Kassahun, M.~Mahmudi, X.~Ricou, R.~Durgesh, A.~S. Chung,
  L.~Hauswald, V.~H. Pham, M.~M{\"{u}}hlegg, S.~Dorn, T.~Fernandez,
  M.~J{\"{a}}nicke, S.~Mirashi, C.~Savani, M.~Sturm, O.~Vorobiov, M.~Oelker,
  S.~Garreis, and P.~Schuberth, ``{A2D2: Audi Autonomous Driving Dataset},''
  apr 2020. [Online]. Available: \url{http://arxiv.org/abs/2004.06320}
\BIBentrySTDinterwordspacing

\bibitem{behley2019iccv}
J.~Behley, M.~Garbade, A.~Milioto, J.~Quenzel, S.~Behnke, C.~Stachniss, and
  J.~Gall, ``{SemanticKITTI: A Dataset for Semantic Scene Understanding of
  LiDAR Sequences},'' in \emph{Proc. of the IEEE/CVF International Conf.~on
  Computer Vision (ICCV)}, 2019.

\bibitem{Goodin2018}
\BIBentryALTinterwordspacing
C.~Goodin, M.~Doude, C.~Hudson, and D.~Carruth, ``{Enabling Off-Road Autonomous
  Navigation-Simulation of LIDAR in Dense Vegetation},'' \emph{Electronics},
  vol.~7, no.~9, p. 154, aug 2018. [Online]. Available:
  \url{http://www.mdpi.com/2079-9292/7/9/154}
\BIBentrySTDinterwordspacing

\bibitem{opencvlibrary}
G.~Bradski, ``{The OpenCV Library},'' \emph{Dr. Dobb's Journal of Software
  Tools}, 2000.

\bibitem{PBPC}
S.~Mishra, G.~Pandey, and S.~Saripalli, ``Extrinsic calibration of a 3d-lidar
  and a camera,'' in \emph{IEEE Intelligent Vehicles Symposium}, 2020.

\bibitem{mishra2020experimental}
S.~Mishra, P.~R. Osteen, G.~Pandey, and S.~Saripalli, ``Experimental evaluation
  of 3d-lidar camera extrinsic calibration,'' in \emph{IEEE International
  Conference on Intelligent Robots and Systems}, 2020.

\bibitem{osteen_2105}
J.~L. {Owens}, P.~R. {Osteen}, and K.~{Daniilidis}, ``Msg-cal: Multi-sensor
  graph-based calibration,'' in \emph{2015 IEEE/RSJ International Conference on
  Intelligent Robots and Systems (IROS)}, 2015, pp. 3660--3667.

\bibitem{Maturana2015}
D.~Maturana and S.~Scherer, ``{VoxNet: A 3D Convolutional Neural Network for
  real-time object recognition},'' in \emph{IEEE Int. Conf. Intell. Robot.
  Syst.}, vol. 2015-Decem.\hskip 1em plus 0.5em minus 0.4em\relax Institute of
  Electrical and Electronics Engineers Inc., dec 2015, pp. 922--928.

\bibitem{Everingham2014}
M.~Everingham, S.~M. Eslami, L.~{Van Gool}, C.~K. Williams, J.~Winn, and
  A.~Zisserman, ``{The Pascal Visual Object Classes Challenge: A
  Retrospective},'' \emph{International Journal of Computer Vision}, vol. 111,
  no.~1, pp. 98--136, jun 2014.

\bibitem{WangSCJDZLMTWLX19}
J.~Wang, K.~Sun, T.~Cheng, B.~Jiang, C.~Deng, Y.~Zhao, D.~Liu, Y.~Mu, M.~Tan,
  X.~Wang, W.~Liu, and B.~Xiao, ``Deep high-resolution representation learning
  for visual recognition,'' \emph{TPAMI}, 2019.

\bibitem{Yuan2019}
\BIBentryALTinterwordspacing
Y.~Yuan, X.~Chen, and J.~Wang, ``{Object-Contextual Representations for
  Semantic Segmentation},'' sep 2019. [Online]. Available:
  \url{http://arxiv.org/abs/1909.11065}
\BIBentrySTDinterwordspacing

\bibitem{Takikawa2019}
\BIBentryALTinterwordspacing
T.~Takikawa, D.~Acuna, V.~Jampani, and S.~Fidler, ``{Gated-SCNN: Gated shape
  CNNs for semantic segmentation},'' in \emph{Proc. IEEE Int. Conf. Comput.
  Vis.}, vol. 2019-October.\hskip 1em plus 0.5em minus 0.4em\relax Institute of
  Electrical and Electronics Engineers Inc., oct 2019, pp. 5228--5237.
  [Online]. Available: \url{http://arxiv.org/abs/1907.05740}
\BIBentrySTDinterwordspacing

\bibitem{Guo2020}
\BIBentryALTinterwordspacing
Y.~Guo, H.~Wang, Q.~Hu, H.~Liu, L.~Liu, and M.~Bennamoun, ``{Deep Learning for
  3D Point Clouds: A Survey},'' \emph{IEEE Transactions on Pattern Analysis and
  Machine Intelligence}, pp. 1--1, jun 2020. [Online]. Available:
  \url{https://github.com/QingyongHu/SoTA-Point-Cloud.}
\BIBentrySTDinterwordspacing

\bibitem{Cortinhal2020}
\BIBentryALTinterwordspacing
T.~Cortinhal, G.~Tzelepis, and E.~E. Aksoy, ``{SalsaNext: Fast,
  Uncertainty-aware Semantic Segmentation of LiDAR Point Clouds for Autonomous
  Driving},'' mar 2020. [Online]. Available:
  \url{http://arxiv.org/abs/2003.03653}
\BIBentrySTDinterwordspacing

\bibitem{Thomas2019}
\BIBentryALTinterwordspacing
H.~Thomas, C.~R. Qi, J.~E. Deschaud, B.~Marcotegui, F.~Goulette, and L.~Guibas,
  ``{KPConv: Flexible and deformable convolution for point clouds},'' in
  \emph{Proc. IEEE Int. Conf. Comput. Vis.}, vol. 2019-October.\hskip 1em plus
  0.5em minus 0.4em\relax Institute of Electrical and Electronics Engineers
  Inc., oct 2019, pp. 6410--6419. [Online]. Available:
  \url{http://arxiv.org/abs/1904.08889}
\BIBentrySTDinterwordspacing

\end{thebibliography}

\end{document}